# A Machine Learning Approach to Detect Dehydration in Afghan Children

Ziaullah Momand, Debajyoti Pal, Pornchai Mongkolnam, Jonathan H. Chan
Innovative Cognitive Computing (IC2) Research Center, School of Information Technology
King Mongkut's University of Technology Thonburi
Bangkok 10140, Thailand
ziaullah.momand1@kmutt.ac.th, debajyoti.pal@sit.kmutt.ac.th, pornchai@sit.kmutt.ac.th, jonathan@sit.kmutt.ac.th

*Abstract*— Child dehydration is a significant health concern, especially among children under 5 years of age who are more susceptible to diarrhea and vomiting. In Afghanistan, severe diarrhea contributes to child mortality due to dehydration. However, there is no evidence of research exploring the potential of machine learning techniques in diagnosing dehydration in Afghan children under five. To fill this gap, this study leveraged various classifiers such as Random Forest, Multilayer Perceptron, Support Vector Machine, J48, and Logistic Regression to develop a predictive model using a dataset of sick children retrieved from the Afghanistan Demographic and Health Survey (ADHS). The primary objective was to determine the dehydration status of children under 5 years. Among all the classifiers, Random Forest proved to be the most effective, achieving an accuracy of 91.46%, precision of 91%, and AUC of 94%. This model can potentially assist healthcare professionals in promptly and accurately identifying dehydration in under five children, leading to timely interventions, and reducing the risk of severe health complications. Our study demonstrates the potential of machine learning techniques in improving the early diagnosis of dehydration in Afghan children.

## I. INTRODUCTION

Dehydration is a condition that occurs when the body loses more fluids than it takes in, resulting in a water deficit that can impair bodily functions. This condition can range from mild to severe and can be caused by various factors, including excessive sweating, diarrhea, vomiting, and not drinking enough fluids. The human body relies on water for maintaining cellular balance, and it makes up approximately 75% of an infant's body weight and up to 60% in adolescents and adults. If people stop consuming water, they would perish within a few days[1].

In a fact, UNICEF estimated approximately 4 billion people, which accounts for almost 2/3 of the world's population, face acute water scarcity for at least one month annually and estimated that by 2040, about 25% of children worldwide will reside in regions with exceedingly high water stress[2]. Dehydration is a common occurrence in children under the age of five, mainly due to the presence of severe diarrhea and vomiting, which are common symptoms of infections such as cholera, rotavirus, and norovirus. Other causes of dehydration in children include inadequate fluid intake, excessive sweating, and high fever [3]. As per a report by the WHO, diarrhea is the second leading cause of death in children under 5 years of age, claiming the lives of approximately 525,000 children annually. The WHO report also highlights that there are nearly 1.7 billion cases of childhood diarrheal disease globally every year [4].

Dehydration is one of the main health issues that is usually ignored by many people, and which can lead to serious health consequences. Dehydration is prevalent in countries with extremely hot and humid weather, where summer temperatures can soar to 50 °c. Nonetheless, it can also occur in colder climates. However, various factors significantly impact a child's dehydration, including lack of water consumption, body temperature, thirst drive, an individual's health status, frequency of urination, and others. Dry mucous membranes, arid axilla, rapid heartbeat, diminished skin elasticity, reduced blood pressure, alterations in urine color, and occasional dizziness are some of the symptoms associated with dehydration [5].

Afghanistan is one of the developing countries that faces a significant challenge of child mortality with a rate of 55.7 per 1000 births as reported by the UNICEF [6]. The reasons for this are multifaceted, with the consequences of civilian wars being one of the primary causes. The country has been in a state of conflict for two decades, which has resulted in a lack of access to basic healthcare services and limited access to humanitarian aid due to security issues. Consequently, there has been a significant impact on the development of the country, and there is a dire need for improved access to safe water and sanitation facilities in villages and towns across the country. A UNICEF report from 2017 highlights that diarrhea-related deaths account for approximately 12% of the annual 80,000 deaths of children under the age of 5 in Afghanistan[7].

Severe acute malnutrition resulting from low body weight can lead to profound dehydration in children under the age of five [8]. In Afghanistan, a country where around 1.2 million children are already malnourished and 41% of children suffer from stunted growth, poor sanitation and hygiene exacerbate the malnourishment and make children more vulnerable to infections that trigger diarrhea, further aggravating their malnourishment [7]. Child dehydration is a critical health concern in Afghanistan, particularly among malnourished children who are already vulnerable due to poor sanitation and hygiene practices. The lack of access to clean drinking water and healthcare facilities further exacerbates the situation, making it difficult to prevent and manage dehydration. As a result, there is a need to develop a machine learning model that can accurately predict and prevent dehydration in Afghan children under 5 years of age, by identifying the most significant risk factors and providing timely interventions to prevent dehydration and its potentially fatal consequences. Recent studies have employed machine learning techniques to determine the hydration status of individuals across various contexts. However, many of these studies rely on invasive sensors to collect data from the human body during clinical trials to evaluate the hydration levels of individuals. In contrast, this study aims to develop a predictive machine learning model that utilizes observational data of sick children to accurately identify dehydration in children under 5 years of age in Afghanistan, without the need for invasive sensors or clinical trials.

The rest of the paper is structured as follows: Related work is discussed first, followed by the method of our work in Section III. Section IV presents the results and Section V provides a detailed discussion. The paper concludes with Section VI, where we draw our final conclusions.

## II. RELATED WORK

The field of machine learning has grown significantly in recent years, with many applications in the healthcare industry. One area where machine learning techniques have been utilized is in the identification of hydration status in individuals undergoing different activities and factors. The ability to accurately predict hydration status has important implications for the prevention of heat-related illness, as well as for optimizing human body performance especially in children. This section aims to provide an overview of the current state of research on the utilization of machine learning techniques for the prediction of hydration status.

Endurance exercise can cause water loss through sweat, managing the hydration level helps regulate body temperature, electrolytes like sodium, potassium, and chloride. Wang et al [9] conducted a study to investigate the use of machine learning models in predicting hydration status during endurance exercise in single-subject experiments. The research involved 32 exercise sessions with and without fluid intake, and four non-invasive physiological and sweat biomarkers were measured: heart rate, core temperature, sweat sodium concentration, and whole-body sweat rate. The authors employed Linear Regression (LR), Support Vector Machine (SVM), and Random Forest (RF) classifiers to predict the percentage of body weight loss as a measure of dehydration using these biomarkers, and the accuracy of the predictions was compared. The findings revealed that the models had similar mean absolute errors, with nonlinear models slightly outperforming linear models. Additionally, the use of whole-body sweat rate or heart rate as biomarkers led to higher prediction accuracy than core temperature or sweat sodium concentration. Similarly, a study was conducted to evaluate and characterize the hydration status and fluid balance characteristics of high-performance adolescent athletes using Linear mixed model (LMM) and K-means cluster techniques [10].

Biochemical sensors are being continuously improved to detect various electrolytes in sweat and evaluate the level of hydration in the human body. Ongoing machine learning research is exploring the use of various sensors to collect body signals and detect dehydration. Liaqat et al [11] used non-invasive wearable sensors and machine learning algorithms to predict hydration levels based on skin conductance data. The RF algorithm achieved the highest accuracy of 91.3% and may offer a new approach for non-invasive hydration detection. A recent study utilized pulse rate variability parameters extracted from photoplethysmography (PPG) signals and electrodermal activity (EDA) to identify mild dehydration. The study evaluated various machine learning techniques, including Linear Discriminant Analysis (LDA), Quadratic Discriminant Analysis (QDA), Logistic Regression, SVM, Gaussian Kernel, K-Nearest Neighbor (KNN), and Decision Tree. The ensemble KNN emerged as the most accurate technique with an accuracy of 91.2% [12]. In a similar vein, Rizwan et al [13] presented a non-invasive auto-detection solution for monitoring the hydration level of individuals using EDA sensors. The study trained six different machine learning algorithms on EDA data and compared their performance using different parameters such as window size and feature combinations. The study found that the K-NN algorithm was the most effective, achieving an accuracy of up to 87.78% for accurately estimating hydration level.

In addition, a context-aware dehydration alert system was developed for non-invasive classification of individuals' hydration level. The study utilized EDA to collect data from the human body and employed various classifiers, including RF, Decision Tree (DT), Naive Bayes (NB), BayesNet, and Multilayer perceptron (MLP), to predict dehydration. The results indicated that the DT classifier was the most effective, achieving an impressive accuracy of 93% [14]. A recent research endeavor explored the utilization of machine learning techniques to monitor hydration levels in athletes, drawing data from wearable sensors that include an accelerometer, magnetometer, gyroscope, galvanic skin response sensor, photoplethysmography sensor, temperature sensor, and barometric pressure sensor. The study authors applied RF and Deep Neural Network models for the purpose of classification [15].

The literature indicates that several machine learning algorithms were employed in previous studies to analyze data, which was collected invasively through human body sensors. While most of these studies did not concentrate on a particular population, a couple of them [9, 14] focused on athletes. However, there is currently no research on the use of machine learning techniques to identify dehydration in children under 5 years of age. Therefore, the objective of this study is to develop a machine learning model based on observational survey data of sick children in Afghanistan, with the aim of identifying dehydrated children under 5 years of age.

## III. METHODOLOGY

The main objective of this study is to develop a predictive model for determining the dehydration status of Afghan children under 5 years of age using machine learning. This section describes the approach we used to achieve this objective, the study began by identifying the problem of dehydration among children in rural areas of Afghanistan. To collect data on this issue, we relied on the Afghanistan Demographic Health Survey (ADHS) conducted between 2018 and 2019, which provided a dataset on sick children in these areas. A detailed account of the steps we took to carry out this is presented in Figure I.

### A. Data Acquisition

When developing a machine learning model, the quality of the data used is paramount. The data must be consistent and derived from a reputable source. In this particular study, the data was obtained from the Demographic Health Survey (DHS) program conducted by the Ministry of Public Health (MoPH) Islamic Republic of Afghanistan with financial aid of USAID in Afghanistan. The DHS program has been collecting various types of data for more than 30 years from different parts of the world [16]. The dataset used in this study is a sample of the Afghanistan Service Provision Assessment (SPA), which is a part of ADHS data focuses on tertiary/specialty and private hospitals. This survey was conducted between November 2018 and January 2019 with the objective of gathering crucial information on the availability, readiness, and quality of health services in Afghanistan.

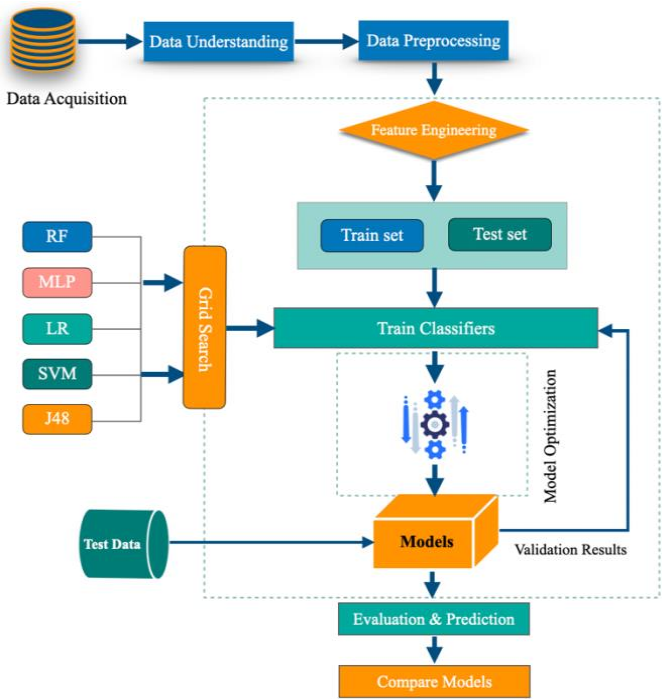

Fig. 1. Diagram illustrating the process of developing a predictive model.

*B. Data Understanding*

In the machine learning pipeline, understanding the data is crucial. We explored the dataset to comprehend its completeness, consistency, variables, and their types. The SPA dataset was organized into 7 distinct categories, which included family planning, facility inventory, staff listing, provider, child health, antenatal care, and normal delivery. Each category was assigned a distinctive code during the data collection process. All variables were then summarized according to their respective unique category code. For this study, child health data was selected from, which comprised 127 attributes with 681 records that provided information about various subjects. The child health data was organized into three general categories: client, observer, and sick child data. The sick child attributes were identified as the primary category of interest, which consisted of 21 attributes. Therefore, 22 attributes related to child illness were extracted from the dataset and are presented in the Table I.

*C. Data Preprocessing*

When training machine learning algorithms, it is important to preprocess the data to improve its quality and usefulness. Preprocessing involves various tasks such as cleaning the data to address missing values, identifying and removing duplicate rows, and identifying and handling outliers and extreme values. In this study, we performed these preprocessing steps to ensure the quality of the data used to train the models. Specifically, we cleaned the data by addressing missing values and duplicate rows, and removed outliers to prevent them from having an undue influence on the model. These preprocessing steps helped to ensure that the machine learning models are trained on high-quality data and can thus produce more accurate and reliable results.

TABLE I. DATASET SHORT DESCRIPTION

| No. | Feature | Description |
|---|---|---|
| 1 | Symptomatic HIV | Child HIV infection |
| 2 | Diarrhea | Child has diarrhea |
| 3 | Respiratory | Child has respiratory disorder |
| 4 | Dysentery | Child has dysentery infection |
| 5 | Amebiasis | Child has infection with amoebas |
| 6 | Malaria | Child has malaria |
| 7 | Fever | Child has fever |
| 8 | Ear | Child has ear problem |
| 9 | Throat | Child has throat problem |
| 10 | Immunized during visit | Child is vaccinated |
| 11 | Age in months | Age of the child in month |
| 12 | Fever in past 2 days | Child had fever in the past 2 days |
| 13 | Cough/breathing | Child has cough/difficulty in breathing |
| 14 | Can drink | Child can drink water |
| 15 | Vomit | Child vomits |
| 16 | Watery frequent stools | Child had watery frequent stools in past 2 days. |
| 17 | Sleepy during illness | Child is excessive sleepy during the ilness |
| 18 | Days ago illness began | Number of days illness began |
| 19 | Nearest facility to home | Health center near to home |
| 20 | Parents are literate | The parents of child are literated |
| 21 | Dehydrated | Child has dehydration |

*D. Feature Engineering*

To improve the quality and relevance of the features used for training the machine learning models, before training the models, feature engineering which included feature selection and transformation was performed. All categorical features were transformed/encoded into the numerical values. The best features based on their relevance were extracted from the total 20 attributes. The attributes evaluation was performed by two feature selection methods. SelectKBest with chi-squared ($\chi^2$) scoring method. It selects the top K features with the highest score from a given set of features based on a user-specified score function. $\chi^2$ is a statistical method used to measure statistical score between each feature and the target variable. This score was given to the SelectKBest method to decide for the K best features. The second method was Mutual Information Gain (MIG) that works on the entropy of variables as shown in (1). To put it simply, MIG refers to the quantity of information that one variable provides regarding the other variable [17].

$$I(X;Y) = H(X) - H(X|Y) \qquad (1)$$

Where $H(X)$ is the marginal entropy, $H(X/Y)$ is the conditional entropy, and $H(X;Y)$ is the joint entropy of X and Y. Table II demonstrates that the features ranked by the MIG method produced the most accurate results for the model, based on experimental analysis. Thirteen features have been identified as the most reliable predictors of dehydration in children under the age of five. It is worth mentioning that both methods used to rank the features were tested by models, but the MIG method yielded more accurate results than another method.

MIG outperforms the Chi-square method in feature extraction due to its information theory basis, ability to capture non-linear relationships, consideration of feature interactions, and robustness to irrelevant features.

*E. Models Implementation*

Once the dataset was preprocessed, the data was ready to feed into the model. Five machine learning algorithms namely, RF, MLP, Logistic Regression (LR), SVM, and J48 classifier were applied to the selected features from the dataset. Due to the presence of categorical and numerical data in the dataset, the aforementioned algorithms can affectively handle the data, particularly, RF, RL, and J48 are able to handle small sample size. During the experimentation, in order to build a robust model, various sub-groups of attributes were fed to the model and the results were evaluated. Although feature selection was performed, certain features were still included in the experiment as they were deemed relevant to dehydration symptoms and causes. Notably, medical literature highlights diarrhea and vomiting as major contributors to dehydration among children below the age of five [18, 19, 20]. Moreover, amebiasis, an infection with amoebas, especially as causing dysentery is also reported as a risk factor dehydration [21]. Therefore, these features were incorporated into the study despite the selection process.

The dataset contained attributes with varying scales, which can pose a challenge for machine learning algorithms. When features have large values, they tend to carry more weight, potentially leading to poorer performance. To address this, the MinMaxScaler technique was applied to normalize all feature values to a range between 0 and 1. This equalizes the weight given to all features, allowing the algorithm to focus on their relative importance.

The dataset was split into 80% training set and 20% testing data. As shown in figure III, the distribution of the classes is unequal and it is skewed towards "No" value of the class. The "Yes" class has a smaller number of examples compared to the "No" class. This poses a challenge for machine learning algorithms since they tend to be biased towards the majority class and produce lower accuracy and recall for the minority class. In order to overcome this problem in the dataset, Synthetic Minority Oversampling Technique (SMOTE) was applied on the training set. SMOTE is a popular approach for dealing with class imbalance problems in machine learning. This technique involves generating synthetic samples for the minority class to create a more balanced dataset. SMOTE is a widely used up-sampling technique that has been shown to improve the performance of machine learning models when working with imbalanced datasets [22]. During the training stage, we conducted model optimization to enhance performance and generalize the capabilities of the models. Hyperparameter tuning was employed to adjust the model's hyperparameters and discover optimal combinations for improved performance. We utilized both Grid Search and Random Search methods with 5-fold cross-validation to identify the best hyperparameters. We compared the run time and performance of both methods.

The Grid Search method achieved high results but had a long runtime. On the other hand, the Random Search method reduced the runtime but had poorer performance. In this case, the Grid Search method was chosen for finding the best hyperparameters because it thoroughly explores all possible combinations of hyperparameter values in a grid. While Random Search randomly samples a subset of the hyperparameter space for a fixed number of iterations.Although Randomized Search is efficient, it doesn't guarantee finding the globally optimal hyperparameters.

TABLE II. BEST FEATURES SELECTED FOR TRAINING THE MODEL

| Rank | Feature | Value |
|---|---|---|
| 1 | Immunized during visit | 0.051948 |
| 2 | Dysentery | 0.043772 |
| 3 | Diarrhea | 0.034325 |
| 4 | Age | 0.024653 |
| 5 | Had watery and frequent stools in the past 2 days | 0.024439 |
| 6 | Child had Fever | 0.010820 |
| 7 | Child can drink | 0.008566 |
| 8 | Excessively sleepy during illness | 0.007740 |
| 9 | Amebiasis | 0.007665 |
| 10 | Days ago illness began | 0.006553 |
| 11 | Child had cough or difficult breathing | 0.006114 |
| 12 | Symptomatic HIV suspected | 0.002117 |
| 13 | Child Vomits | 0.002113 |

*F. Model Evaluation*

Evaluation matrices namely, accuracy, precision, recall, and Area Under the ROC Curve (AUC) were considered to evaluate the model performance.

*1) Accuracy:*

Typically, accuracy is the most widely employed metric, which calculates the proportion of correctly classified instances out of the total evaluated instances as shown in (2).

$$Accuracy = \frac{TP+TN}{TP+FP+TN+FN} \quad (2)$$

Where TP represents the number of correctly predicted positive instances, while TN represents the number of correctly predicted negative instances. FP are instances that are predicted to be positive but are actually negative, while FN are instances that are predicted to be negative but are actually positive.

*2) Precision*

Used for binary classification that measures how often the model's positive predictions are actually correct as presented in (3).

$$Precision = \frac{TP}{TP+FP} \quad (3)$$

*3) Recall*

A performance matrix for binary classification that calculates how often the model correctly identifies positive instances in the dataset as shown in the (4).

$$Recall = \frac{TP}{TP+FN} \quad (4)$$

*4) F1-Score*

The F1-score, derived from precision ($p$) and recall ($r$), is a concise metric that evaluates a model's binary classification performance. It provides a balanced assessment, ranging from 0 to 1, where higher values indicate superior overall performance as shown in (5).

$$F1-score = 2 * \frac{p*r}{p+r} \quad (5)$$

*5) Area Under the ROC Curve (AUC)*

It measures the model's ability to distinguish between positive and negative classes by plotting the True Positive Rate against the False Positive Rate at various threshold values. The AUC represents the overall performance of the model, where a value of 1 represents a perfect classifier, and a value of 0.5 represents a random classifier as shown in (6).

$$AUC = \frac{s_p - n_p(n_n+1)/2}{n_p n_n} \quad (6)$$

In this context, $s_p$ refers to the sum of the ranks of all positive examples, while $n_p$ and $n_n$ represent the respective number of positive and negative examples in the dataset [23].

## IV. RESULTS

This section presents the experimentation results aimed at identifying dehydration in children. To achieve this goal, five machine learning classifiers namely RF, MLP, LR, SVM, and J48 were used and the performance of the models were compared as shown in Table III.

Table III shows that the RF classifier demonstrated strong performance across several evaluation metrics, including an accuracy of 91.46%, precision of 91%, and AUC of 94%. In comparison, the MLP classifier achieved an accuracy of 85.48%, precision of 85%, recall of 85%, and AUC of 76%. Although the SVM obtained a considerable AUC of 83% after the RF classifier, it fell short in other metrics. The LR classifier also obtained good AUC of 79.53%. However, the LR including J48 classifiers did not perform well in other evaluation metrics. Overall, the RF classifier was the strongest performer in the study, with a notable advantage in terms of its AUC score. As shown in Figure II, the RF AUC score of 94% indicates that the classifier has a high ability to distinguish between positive and negative samples. A perfect classifier would have an AUC of 1, indicating perfect discrimination between positive and negative classes, while a random guess classifier would have an AUC of 0.5.

## V. DISCUSSION

The results obtained from the 5 classifiers, as presented in Table III, demonstrate significant variability across all evaluation metrics. Notably, the RF classifier achieved outstanding performance in both accuracy and AUC metrics. However, due to the initial unbalanced nature of the dataset, the classifiers initially produced poor results. This issue was addressed by applying the SMOTE resampling method to the training dataset, resulting in considerable improvements, particularly for the RF, MLP, and J48 classifiers.

The MLP classifier outperformed in comparison with 3 other classifiers with an accuracy of 85.48%, achieved using best hyperparameters defined by grid search. However, the AUC metric of the MLP classifier was lower compared to its accuracy metric. This could be attributed to the fact that MLP classifiers require large amounts of training data to learn complex patterns, and in this case, the dataset may have been insufficient. SVM is capable of finding non-linear decision boundaries between the positive and negative classes. Thus, SVM achieved high AUC score in comparison with MLP, LR, and J48. However, the high accuracy and AUC score achieved by the RF classifier suggests that the RF algorithm could be a more effective classifier for identifying the dehydration status of Afghan children under 5 years of age, compared to the other classifiers evaluated in the study. The strong performance across multiple evaluation metrics, including accuracy and precision, further supports the potential effectiveness of the RF classifier for this specific application.

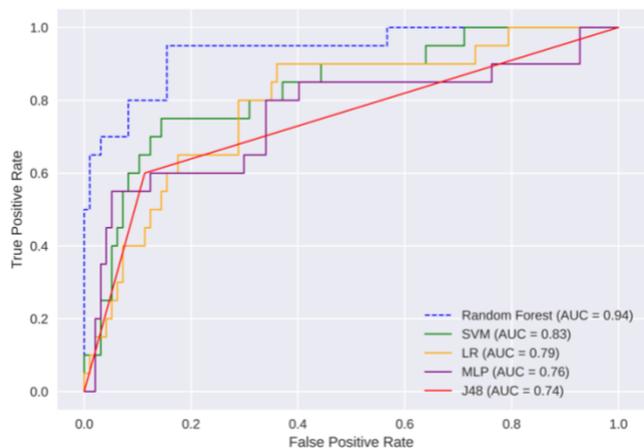

Fig. 2. ROC curve for RF, SVM, LR, MLP, and J48 classifiers.

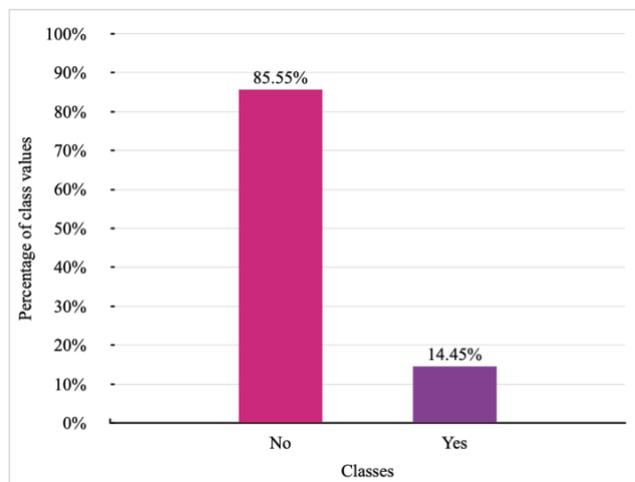

Fig. 3. The distribution of majority and minority classes in the dataset.

TABLE III. RESULTS FROM FIVE CLASSIFIERS

| Classifier | Accuracy | Precsion | Recall | F1-Score | AUC |
|---|---|---|---|---|---|
| RF | 91.46% | 91.00% | 91.00% | 91.00% | 94.00% |
| MLP | 85.48% | 85.00% | 85.00% | 85.00% | 76.00% |
| J48 | 84.56% | 88.00% | 85.00% | 86.00% | 74.33% |
| SVM | 82.05% | 82.00% | 82.00% | 82.00% | 83.00% |
| LR | 70.95% | 85.00% | 71.00% | 74.00% | 79.53% |

This study aimed to identify the dehydration status in Afghan children under 5 years of age. The dataset used in this study was the real observational data collected by MoPH of Afghanistan from hospitals. The existing studies [9, 10, 11, 13, 14, 15] have utilized skin wearable sensors data to train dehydration identification models, which require invasive sensors that can cause discomfort and potential harm to the wearer, including skin irritation or damage.

This study utilized real observational data of sick children that included general symptoms of dehydration and was validated with medical literature, to build a dehydration identification model for children under 5 years of age in the context of Afghanistan. This approach eliminates the need for invasive sensors and provides a non-intrusive and practical method for identifying dehydration in this vulnerable population.

The potential applications of machine learning in assessing children's dehydration status have not been investigated in previous research studies. This study has the potential to be a valuable initiative in drawing attention to the need for improved diagnosis of dehydration in children, particularly in Afghanistan. By exploring the applications of machine learning in this area, researchers may gain new insights and develop more accurate diagnostic tools for detecting and addressing dehydration in children under 5 years of age. However, one important limitation of this study is the relatively small sample size used, which may limit the generalizability of the findings. The dataset used was also limited in size and scope, which may have impacted the accuracy of the results obtained. Future studies with larger, more diverse datasets may be needed to further validate the efficacy of machine learning in diagnosing dehydration in children under 5 years of age.

## VI. Conclusion

This study represents a valuable step towards building a predictive model for predicting dehydration in children under 5 years of age, particularly in resource-limited settings like Afghanistan. Five machine learning classifiers were applied to a sample of Afghan sick children dataset. Random Forest classifier demonstrated outstanding results. By exploring the potential applications of machine learning in this context, our study has demonstrated the feasibility and potential benefits of using this technology to improve clinical outcomes for children suffering from dehydration. This study incorporated a relatively small dataset. Further research is needed to validate the findings of this study and to explore the implications of using this model for diagnostic purposes in vulnerable populations such as children in a large set of data. This study highlights the potential of machine learning to improve clinical practice and underscores the importance of continued research in this area to improve health outcomes for under 5 children with dehydration diagnosis.